%% file: main.tex
\documentclass[10pt,twocolumn,letterpaper]{article}

\usepackage{cvpr}              
\input{preamble}
\definecolor{cvprblue}{rgb}{0.21,0.49,0.74}
\usepackage[accsupp]{axessibility}  
\usepackage[pagebackref,breaklinks,colorlinks,allcolors=cvprblue]{hyperref}
\usepackage{booktabs}
\usepackage{capt-of} 
\usepackage{multirow}
\usepackage{booktabs}  
\usepackage{arydshln}  
\usepackage{makecell}  
\usepackage{tikz}       

\def\paperID{11049} 
\def\confName{CVPR}
\def\confYear{2026}

\title{AnchorSplat: Feed-Forward 3D Gaussian Splatting with 3D Geometric Priors}

\def\authors{
    Xiaoxue Zhang \quad 
    Xiaoxu Zheng \quad
    Yixuan Yin \quad
    Tiao Zhao \quad
    Kaihua Tang \\
    Michael Bi Mi\quad
    Zhan Xu \quad
    Dave Zhenyu Chen$^{{*}}$ \quad
}

\author{
    \authors \\
    \textit{Huawei Technologies Ltd.}
}

\begin{document}

\input{figures/teaser}

\footnotetext{\small $^{{*}}${Corresponding Author}}

\input{sec/0_abstract}    
\input{sec/1_intro}

\input{sec/2_relatedworks}

\input{sec/3_method}
\input{sec/4_experiments}
\input{sec/5_conclusion}

\clearpage

{
    \small
    \bibliographystyle{ieeenat_fullname}
    \bibliography{main}
}

\input{sec/X_suppl}

\end{document}

%% file: figures/teaser.tex
\begin{figure}
    \captionsetup{singlelinecheck=false}
    \twocolumn[{
        \renewcommand\twocolumn[1][]{#1}
        \maketitle
        \vspace{-5mm}
        \centering
        \setlength\tabcolsep{0pt}
        \includegraphics[width=\linewidth]{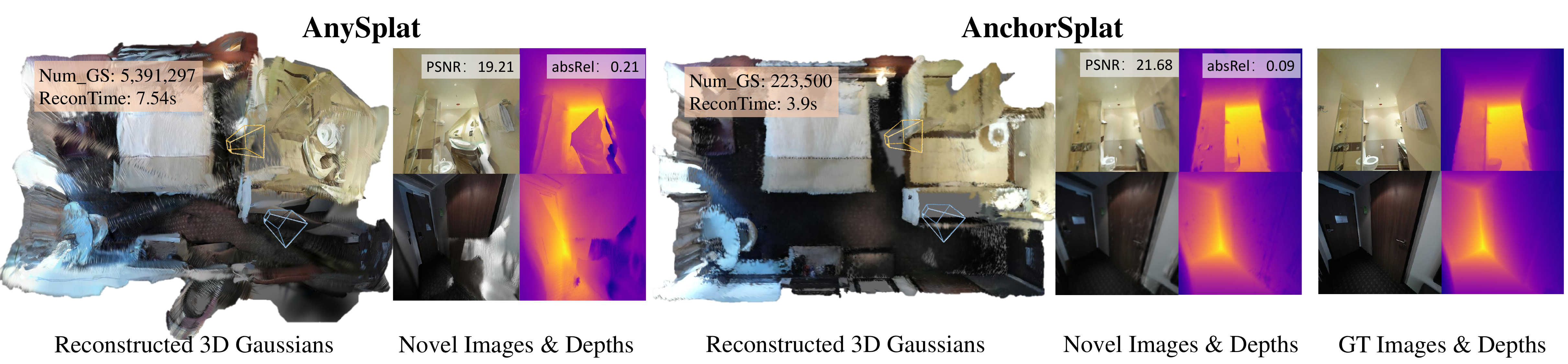}
        \caption{
            Novel view synthesis comparison between AnySplat~\cite{jiang2025anysplat} and our AnchorSplat. AnySplat, as a voxel-aligned feed-forward approach, often suffers from inaccurate geometry, leading to ghosting artifacts, floaters, and blurred structures, especially in regions with sparse or ambiguous depth cues. In contrast, given the same sequence of input images, our anchor-aligned approach performs feed-forward Gaussian reconstruction using nearly 20× fewer Gaussians and about half the reconstruction time, while producing sharper geometry, cleaner surfaces, artifact-free renderings, and more accurate depth estimation.
        }
        \vspace{4mm}
        \label{fig:teaser}
    }]
\end{figure}

%% file: sec/0_abstract.tex
\begin{abstract}
Recent feed-forward Gaussian reconstruction models adopt a pixel-aligned formulation that maps each 2D pixel to a 3D Gaussian, entangling Gaussian representations tightly with the input images. 
In this paper, we propose AnchorSplat, a novel feed-forward 3DGS framework for scene-level reconstruction that represents the scene directly in 3D space. 
AnchorSplat introduces an anchor-aligned Gaussian representation guided by 3D geometric priors (e.g., sparse point clouds, voxels, or RGB-D point clouds), enabling a more geometry-aware renderable 3D Gaussians that is independent of image resolution and number of views. 
This design substantially reduces the number of required Gaussians, improving computational efficiency while enhancing reconstruction fidelity. 
Beyond the anchor-aligned design, we utilize a Gaussian Refiner to adjust the intermediate Gaussians via merely a few forward passes.
Experiments on the ScanNet++ v2 NVS benchmark demonstrate the SOTA performance, outperforming previous methods with more view-consistent and substantially fewer Gaussian primitives. 
\end{abstract}

%% file: sec/1_intro.tex
\section{Introduction}
\label{sec:intro}

Scene-level 3D reconstruction~\cite{Schonberger_2016_CVPR,kerbl20233d,mildenhall2020nerf} is a fundamental problem in computer vision, with broad downstream applications in robotics~\cite{zhang2023part,han2022scene}, augmented reality~\cite{watson2023heightfields,cao2020accurate}, and autonomous navigation~\cite{lei2025gaussnav,chen2025splat}. Existing optimization-based approaches, such as 3D Gaussian Splatting (3DGS)~\cite{kerbl20233d,chen2025survey3dgs} and Neural Radiance Fields (NeRF)~\cite{mildenhall2020nerf,gao2022nerf}, achieve high-fidelity 3D reconstruction, yet they are computationally expensive due to their per-scene iterative optimization. In addition, their depth estimation accuracy and geometric consistency are often suboptimal, restricting their applicability to real-world scenarios.

To alleviate the computational overhead introduced by per-scene iterative optimization, recent feed-forward 3DGS methods directly predict pixel-aligned Gaussians to enable cross-scene generalization~\cite{Charatan_2024_CVPR,chen2024mvsplat,xu2025depthsplat}. While these methods significantly improve efficiency, they also introduce several inherent limitations: (1) reconstruction quality is heavily dependent on the reliability of predicted depth maps or cost volumes, and since back-projected pixel features interact weakly with neighboring points in 3D space, these methods often produce floaters, fragmented surfaces, and inconsistent geometry; (2) the representation remains bound to the 2D grid, \textit{i.e.}, $V \times H \times W$, which leads to redundant Gaussian primitives in plain regions and insufficient coverage in geometrically complex areas; (3) pixel-aligned formulations are sensitive to occlusions, low-texture regions, and motion parallax, resulting in inconsistent sampling patterns across views, as shown in Fig.~\ref{fig:comp_representation}; (4) most existing methods only interpolate within limited view ranges, restricting their ability to reconstruct large-scale environments or extrapolate to unseen viewpoints.

To address these limitations, we propose \textbf{AnchorSplat}, a feed-forward, anchor-aligned 3DGS framework that learns scene representations directly in 3D space from anchor priors. Specifically, we back-project multi-view image features into an anchored feature volume, forming a consistent 3D prior for Gaussian prediction. This anchor-aligned design removes the dependence on reference views and expensive volume computation, enabling accurate depth and geometry prediction even under sparse-view inputs.
Our framework consists of a Gaussian decoder and a plug-and-play Gaussian refiner. The refiner can be applied independently to improve Gaussian attributes without retraining the full model. By aggregating multi-view features into coherent 3D anchors before Gaussian prediction, AnchorSplat reduces floaters and cross-view inconsistencies, resulting in high-fidelity and geometrically consistent 3D reconstructions. Moreover, the anchor-aligned formulation naturally supports the integration of auxiliary 3D signals, such as point clouds or voxel grids. The resulting high-precision 3DGS representation also provides robust coordinates for downstream tasks, including scene understanding~\cite{zhou2024hugs,shi2024language}, navigation~\cite{chen2025splat,honda2025gsplatvnm}, and 3D reasoning~\cite{shi2024language,liu2025reasongrounder,qiu2026slarm}.

In our experiments, we conduct extensive evaluations on the ScanNet++ V2~\cite{yeshwanth2023scannet++} benchmark. Across this dataset, AnchorSplat consistently outperforms existing approaches in reconstruction fidelity, geometric consistency, and view generalization. Notably, AnchorSplat achieves these improvements while using significantly fewer Gaussians and requiring less reconstruction time, highlighting both its efficiency and scalability. These results demonstrate that AnchorSplat is a practical and highly generalizable solution for real-world, scene-level 3D reconstruction.

The contributions of this paper are threefold:

\begin{itemize}
\item \textbf{An anchor-aligned feed-forward Gaussian reconstruction model.} We introduce an anchor-aligned feed-forward Gaussian reconstruction model framework that provides a practical end-to-end solution for scale-consistent depth prediction and arbitrary-view rendering.
\item \textbf{A plug-and-play Gaussian refiner.} We propose a plug-and-play Gaussian refiner module that enhances Gaussian attributes and improves geometric consistency without retraining the entire network.
\item \textbf{Superior performance with higher efficiency.} AnchorSplat delivers superior reconstruction quality while using fewer Gaussians and significantly reducing reconstruction time.
\end{itemize}

%% file: sec/2_relatedworks.tex
\section{Related Works}
\label{sec:related_works}

\input{figures/PixelvsAnchor}

\noindent\textbf{Per-scene neural rendering.} Early NeRF-based methods~\cite{mildenhall2020nerf,niemeyer2022regnerf,yang2023freenerf} achieve high-quality novel view synthesis from calibrated dense multi-view images, but require compute-heavy per-scene optimization and accurate camera parameters from Structure-from-Motion systems~\cite{Schonberger_2016_CVPR}.
3D Gaussian Splatting (3DGS)~\cite{kerbl20233d} offers a faster alternative. The original 3DGS optimizes Gaussian positions, scales, colors, and opacities per scene, achieving orders-of-magnitude speedups over NeRF while maintaining comparable quality. 
Follow-up work extend optimization-based 3DGS in different ways, including reducing reconstruction time~\cite{mallick2024taming,chen2025dashgaussian}, addressing aliasing artifacts~\cite{yu2024mip}, and handling large-scale~\cite{lin2024vastgaussian} or sparse-view inputs~\cite{zhu2025fsgs,zhang2024cor,zhong2025taming,chen2025quantifying}.

\noindent\textbf{Feed-forward 3D reconstruction models.}
Recent work focuses on feed-forward 3D reconstruction from unposed image collections, leveraging geometry and camera poses as a competitive substitute for SfM initializations in NeRF and 3DGS. The pioneering work~\cite{wang2024dust3r} directly regresses dense point maps and relative camera parameters from image pairs without SfM, while subsequent methods~\cite{yang2025fast3r,wang2025continuous,wang2025vggt} extend this approach to hundreds of images in a single transformer-based forward pass, eliminating pairwise matching and iterative alignment. MapAnything~\cite{keetha2025mapanything} introduces a factored multi-view representation—comprising local depth maps, ray confidences, and a global scale—trained across diverse input settings.
The LRM family, including LRM~\cite{hong2023lrm}, GS-LRM~\cite{zhang2024gs}, and iLRM~\cite{kang2025ilrm}, employs latent scene representations to efficiently encode geometry across diverse scenes, providing complementary insights into scalable 3D scene encoding. More recent feed-forward methods, such as YoNoSplat~\cite{ye2025yonosplat}, WorldMirror~\cite{liu2025worldmirror}, and VolSplat~\cite{wang2025volsplat}, further improve reconstruction efficiency and fidelity.
These feed-forward reconstruction pipelines can either replace traditional SfM/MVS or provide reliable poses, depth maps, and point clouds as geometric priors. Our method follows this trend by leveraging such priors to construct an anchor-aligned Gaussian representation for novel view synthesis.

\noindent\textbf{Generalizable 3DGS.}
Instead of optimizing 3DGS per scene, generalizable 3DGS networks predict scene representations in a single forward pass. 
A common strategy is to predict Gaussians binded with input pixels~\cite{Szymanowicz2023SplatterIU,Charatan_2024_CVPR,wewer2024latentsplat,chen2024mvsplat,wang2024freesplat,liu2024mvsgaussian,xu2025depthsplat,xu2025resplat}.
Albeit generalizable, the number and distribution of their Gaussians remain tightly coupled to image resolution and viewpoint coverage, leading to a linear growth of total Gaussians according to the number of input views. 
AnySplat~\cite{jiang2025anysplat} represents a recent breakthrough in pose-free feed-forward 3DGS framework by aggregating redundant pixel-aligned Gaussians via a differentiable voxelization module.
However, as the final representation is still originated from intermediate pixel-aligned Gaussians, it remains sensitive to image resolutions and viewpoint coverages. 
In contrast, our framework starts directly from 3D anchor points from geometric priors, producing anchor-aligned Gaussians that are independent of input pixels, significantly reducing the number of Gaussians and improving cross-view alignment.

%% file: figures/PixelvsAnchor.tex
\begin{figure}[tp]
  \centering
   \includegraphics[width=\linewidth, trim=6 70 2 70, clip]{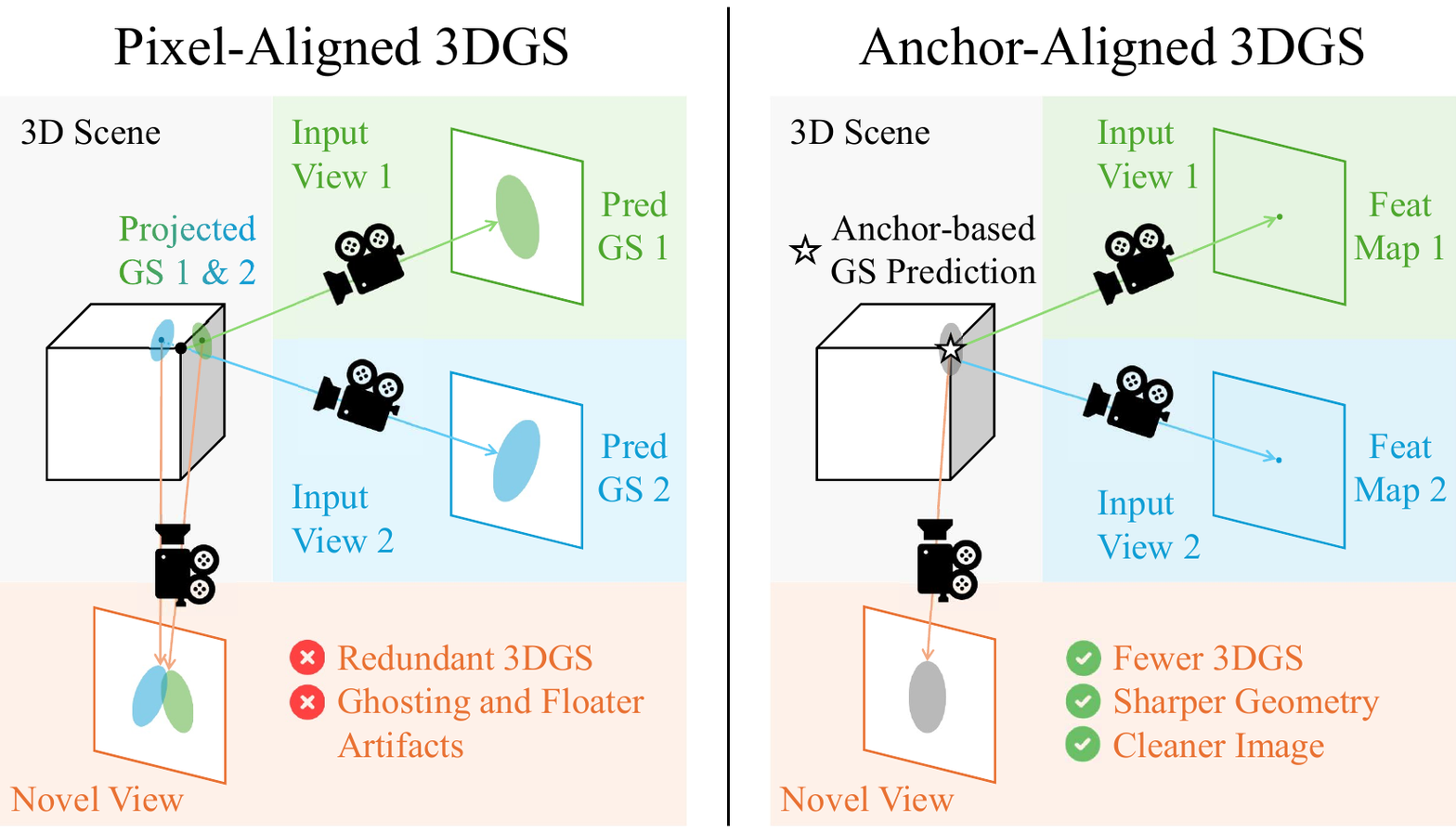}
   \caption{Comparison of pixel-aligned and anchor-aligned Gaussian representations. Pixel-aligned Gaussians exhibit inconsistent sampling across views, especially under occlusions, low-texture regions, and motion parallax, while anchor-aligned Gaussians provide a more stable and consistent 3D representation.}
   \label{fig:comp_representation}
   \vspace{-0.5cm}
\end{figure}

%% file: sec/3_method.tex
\section{Method}
\label{sec:method}

\input{figures/pipeline}

\subsection{Preliminary}
\label{subsec:preliminary}

Feed-forward 3D reconstruction aims to learn a mapping from a set of posed or unposed input images $V$ to a 3D representation, such as 3D Gaussians, meshes, or radiance fields. Pixel-aligned 3D Gaussian Splatting (3DGS) methods extract image features and refine them via cross-view interactions:
\begin{equation}
    \{F_i\}_{i=1}^V = f\left(\theta\left(\{I_i, P_i\}_{i=1}^V\right)\right),
\end{equation}
where $F_i\in \mathbb R^{3 \times h \times w}$, $\theta$ denotes a pretrained image encoder, and $f$ is a feature aggregation function that fuses multi-view information (e.g., feature matching and fusion). For pixel-aligned 3D reconstruction, these features are upsampled to the input image resolution, producing the output Gaussians
\begin{equation}
    \{\mathcal{G}\}_{i=1}^N = \phi(\{F_i\}_{i=1}^V),
\end{equation}
with $N = H \times W \times V$. Thus, each pixel corresponds to a unique Gaussian, and the total number of Gaussians equals the total number of pixels across all views.  
However, pixel-aligned methods have several limitations:  
1) The 3D Gaussians are tightly coupled to the 2D pixels, resulting in view-biased Gaussian densities—regions frequently observed across views generate more Gaussians, whereas texture-rich or structurally complex areas may be underrepresented.  
2) Feature extraction and cross-view interactions occur in 2D image space, limiting interactions among neighboring points in 3D space and causing floating artifacts.  
3) Geometric consistency is difficult to guarantee, and redundant Gaussians often lead to blur.  
To address these issues, we propose an \textit{anchor-aligned 3D representation} that is sparse, consistent, and naturally distributed in 3D space.

\subsection{Problem Setup}
Given $N$ images $\{I_i\}_{i=1}^V \in \mathbb{R}^{H \times W \times 3}$, AnchorSplat aims to learn a feed-forward model that simultaneously reconstructs 3D geometry and appearance using 3D Gaussian Splatting. Each 3D Gaussian is represented as $\mathcal{G} = \{\mu, \alpha, \Sigma, sh\}$,
where $\mu \in \mathbb{R}^3$ is the Gaussian center, $\alpha \in \mathbb{R}$ denotes opacity, $\Sigma \in \mathbb{R}^7$ encodes covariance, including scale $s \in \mathbb{R}^3$ and rotation $r \in \mathbb{R}^4$, and $sh \in \mathbb{R}^{3(\text{deg}+1)^2}$ are the spherical harmonics coefficients representing colors. Therefore, the reconstruction can be formally expressed as:
\begin{equation}
    \{\mathcal{G}_g\}_{g=1}^{N} = \mathrm{AnchorSplat}(\{I_i\}_{i=1}^V).
\end{equation}

\subsection{Anchor-aligned 3DGS Reconstruction}
As shown in Fig.~\ref{fig:Pipeline}, AnchorSplat consists of three main components: an anchor predictor, a Gaussian decoder, and a Gaussian refiner.  
First, a pretrained Multi-View Stereo module (such as MapAnything~\cite{keetha2025mapanything}, MVSAnywhere~\cite{izquierdo2025mvsanywhere}, Reliev3R~\cite{chen2026reliev3r}, etc) predicts reliable 3D geometric priors and depth maps from the input images, which serve as spatial anchors for the following stages. Next, the posed images and depths are processed by a CNN-based feature extractor to encode multi-view information. These features are then fed into a transformer-based Gaussian decoder to predict anchor-aligned Gaussians consistent with the 3D priors. Finally, the Gaussian refiner adjusts and enhances the predicted Gaussian attributes (position, scale, opacity, and color) to improve reconstruction fidelity and view consistency.

\noindent\textbf{Anchor Predictor.}
We leverage the pretrained MapAnything~\cite{keetha2025mapanything}, which is one of effective multi-view stereo models, to predict depths and camera poses from the $V$ unposed input images. These predictions are then back-projected into 3D space to obtain the corresponding 3D geometry points. However, the resulting points are still pixel-aligned to the resolution of the input images; that is, $V$ images of resolution $H \times W \times 3$ produce $V \times H \times W$ 3D points. As discussed in Section~\ref{subsec:preliminary}, this pixel-aligned representation introduces significant computational redundancy, since each pixel of every view generates a separate 3D Gaussian, many of which occupy overlapping regions in 3D space.

To address this, the predicted pixel-aligned 3D points are downsampled into a sparser set of \textit{anchors} by using farthest point sampling (FPS) algorithm, whose downsampling number is determined by voxelizing the 3D space:
\begin{equation}
\begin{aligned}
\left\{D_i, K_i, P_i \right\}_{i=1}^{V} &= \mathrm{MapAnything}(\{I_i\}_{i=1}^V), \\
\{A_j\}_{j=1}^{N} &= \mathrm{ds}\Big(\mathrm{Proj}(\{D_i, K_i, P_i\}_{i=1}^{V})\Big),
\end{aligned}   
\end{equation}
where $\mathrm{ds}$ denotes the downsampling operation (here, we utilize the FPS algorithm), $D_i, K_i, P_i$ are the depth, intrinsic, and extrinsic parameters of view $i$, and $A_j \in \mathbb{R}^3$ represents the 3D anchors with $N \ll H \times W \times V$.  

Specifically, a 2D depth $D_i(u,v)$ from view $i$ can be lifted to 3D world coordinates $P_w$ using the corresponding camera intrinsics $K_i$ and extrinsics $(R_i, T_i)$:
\begin{equation}
\label{eq:proj}
    P_w = R_i P_c + T_i = R_i \Big( D_i(u,v) K_i^{-1} \begin{bmatrix} u & v & 1 \end{bmatrix}^\top \Big) + T_i.
\end{equation}
This process converts dense pixel-aligned points into a sparse set of anchors, reducing computational cost while preserving geometric fidelity.

\noindent\textbf{Gaussian Decoder.}
Given the 3D geometry anchors $\{A_j\}_{j=1}^N$, along with the depths $\{D_i\}_{i=1}^V$ and camera poses $\{K_i, R_i, T_i\}_{i=1}^V$, we first extract multi-view features for each anchor. To do this, we use a lightweight 2D U-Net to encode the input images $\{I_i\}$, corresponding depth maps $\{D_i\}$, and camera ray embeddings $\mathrm{Ray}(K_i, R_i, T_i)$:  
\begin{equation}
    F_i = E(I_i, D_i, \mathrm{Ray}_i), \quad \forall i \in \{1, \dots, V\},
\end{equation}
where $(I_i, D_i, \mathrm{Ray}_i) \in \mathbb{R}^{H \times W \times (3 + 1 + 6)}$, and $F_i \in \mathbb{R}^{h \times w \times C}$, with $C$ denoting the feature dimension.

Next, the 2D features $\{F_i\}_{i=1}^V$ are projected onto the 3D geometry anchors $\{A_j\}_{j=1}^N$ using the depth and camera poses via Eq.~\ref{eq:proj}, producing anchor features $\{\tilde{A}_j\}_{j=1}^N \in \mathbb{R}^{C}$.  
These features are then processed by a transformer-based Gaussian predictor to model 3D spatial interactions among all anchors. An MLP subsequently predicts Gaussian attributes, including the offset of the center $\delta \mu$, opacity $\alpha$, scale $s$, rotation $r$, and spherical harmonic coefficients $sh$. Each anchor predicts four Gaussians:  
\begin{gather}
    \{\mathcal F_j\}_{j=1}^N = \mathrm{Attn}(\{F_j, A_j\}_{j=1}^N), \\
    \{\delta \mu_j, \alpha_j, s_j, r_j, sh_j\}_{j=1}^{4N} = \mathrm{MLP}(\{\mathcal F_j\}_{j=1}^N).
\end{gather}

Finally, the absolute Gaussian centers are obtained as $\mu_j = A_j + \delta \mu_j$,
where the predicted Gaussians are typically constrained to lie within a small range (e.g., $10/128$) around their corresponding anchor positions $A_j$.

\noindent\textbf{Gaussian Refiner.}
Due to the limited number of anchors and Gaussians, as well as the constraints of a generalized feed-forward architecture, some regions in the rendered images may appear blurred or contain holes, indicating that the Gaussian attributes still have room for improvement. To address this, we design a Gaussian Refiner module to further enhance the reconstruction quality based on the Gaussians predicted by the Gaussian Decoder.
In this module, the rendering error guides the update of Gaussian attributes. Specifically, for each input view $i$, the difference between the rendered image $\hat I_i$ and the ground-truth image $I_i$, together with the current Gaussian attributes $\mathcal{G}_j$ and the aggregated anchor features $\tilde F_j$, serves as input to the refiner.  

Inspired by G3R~\cite{chen2024g3r}, a pretrained ResNet-18~\cite{he2016deep} extracts multi-scale features (1/2, 1/4,  1/8) from both $\hat I_i$ and $I_i$. The features are resized and concatenated to a common 1/4 resolution, resulting in rendered features $\hat F_i$ and ground-truth features $F_i$ for each view $i$, with $\hat F_i, F_i \in \mathbb{R}^{\frac{H}{4} \times \frac{W}{4} \times D}$
where $D$ denotes the feature dimension. The per-view rendering error is then computed as $e_i = F_i - \hat F_i, \quad \forall i \in \{1, \dots, V\}$
A differentiable back-projection lifts the 2D render errors $\{e_i\}_{i=1}^V$ to the corresponding 3D Gaussians using the depth maps and camera poses:
\begin{equation}
    \left\{E_j \in \mathbb{R}^D\right\}_{j=1}^{4N} = \mathrm{agg} \Big( \mathrm{proj}\left(\left\{e_i, K_i, (R_i, T_i), D_i\right\}_{i=1}^V\right) \Big),
\end{equation}
where $\mathrm{proj}$ maps 2D errors to 3D Gaussian locations and $\mathrm{agg}$ aggregates multi-view contributions.

Next, a transformer block captures the spatial interactions among the rendering errors of all Gaussians:
\begin{equation}
   \left\{ \mathcal{E}_j \right\}_{j=1}^{4N} = \mathrm{Attn}\!\left( \left\{ E_j \right\}_{j=1}^{4N} \right).
\end{equation}
Finally, a Point Transformer~\cite{wu2024point} updates the Gaussian attributes using the combination of current attributes $\mathcal{G}_j$, anchor features $\hat{\mathcal{F}}_j$, and the refined error features $\mathcal{E}_j$:
\begin{equation}
\left\{ \delta \mathcal{G}_j \right\}_{j=1}^{4N}
 = \mathrm{SerialAttn}\!\left( 
     \left\{ \mathcal{G}_j,\, \hat{\mathcal{F}}_j,\, \mathcal{E}_j \right\}_{j=1}^{4N} 
 \right),
\end{equation}
where $\hat{\mathcal{F}}_j := \mathcal{F}_{\,p(j)}, \quad
    p(j) = \Big\lceil \frac{j}{4} \Big\rceil, \quad j = 1, \dots, 4N, \nonumber$
i.e., each anchor feature $\mathcal{F}_i$ is shared across its four corresponding Gaussians. The final updated Gaussian attributes are 
\begin{equation}
    \hat{\mathcal{G}}_j = \mathcal{G}_j + \delta \mathcal{G}_j.
\end{equation}

\subsection{Training Objective}
Our model is trained in two stages. In the first stage, we train the Gaussian Decoder to predict Gaussians grown from the anchors provided by the Anchor Predictor module. The training loss is a combination of a rendering loss $\ell_I$, a depth loss $\ell_D$, and regularization terms on the Gaussian opacity $\ell_{\alpha}$ and volume $\ell_s$:
\begin{align}
    L_{GSdec} =& \lambda_I \sum_{i=1}^{V} \ell_{I}(\hat I_i, I_i) + \lambda_D \sum_{i=1}^{V} \ell_{1} (\hat D_i, D_i) + \nonumber\\
    &\lambda_{\alpha} \ell_{\alpha} (\alpha_j) + \lambda_{s} \ell_s  (s_j), \nonumber\\
    \ell_I (\hat I_i, I_i) =& \ell_1(\hat I_i, I_i) + \gamma_{\mathrm{SSIM}}(1-\mathrm{SSIM}(\hat I_i, I_i)) + \nonumber \\
    &\gamma_{\mathrm{LPIPS}} \mathrm{LPIPS}(\hat I_i, I_i) \nonumber\\ 
    \ell_\alpha (\alpha_j) =& \frac{1}{4N}\sum_{j=1}^{4N} \left( 1-\alpha_j \right), \nonumber \\
    \ell_s (s_j) =& \frac{1}{4N}\sum_{j=1}^{4N} \prod s_j,
\end{align}
where $V$ is the number of views rendered per training step, and $\alpha_j$ and $s_j$ denote the opacity and scale of Gaussian $j$, respectively. The terms $\ell_\alpha$ and $\ell_s$ prevent Gaussians from becoming overly transparent or excessively large. In our experiments, the loss weights are set to $\lambda_I=200$, $\gamma_\mathrm{SSIM}=0.2$, $\gamma_\mathrm{LPIPS}=0.2$, $\lambda_D=100$, $\lambda_\alpha=1\mathrm{e}{-1}$, and $\lambda_s=1\mathrm{e}{4}$. In the second stage, we freeze the Gaussian Decoder and train only the Gaussian Refiner. During this stage, only the rendering loss $\ell_I(\tilde I_i, I_i)$ is applied to supervise the rendered images against the ground-truth images.

%% file: figures/pipeline.tex
\begin{figure*}[tp]
  \centering
  \includegraphics[width=\linewidth]{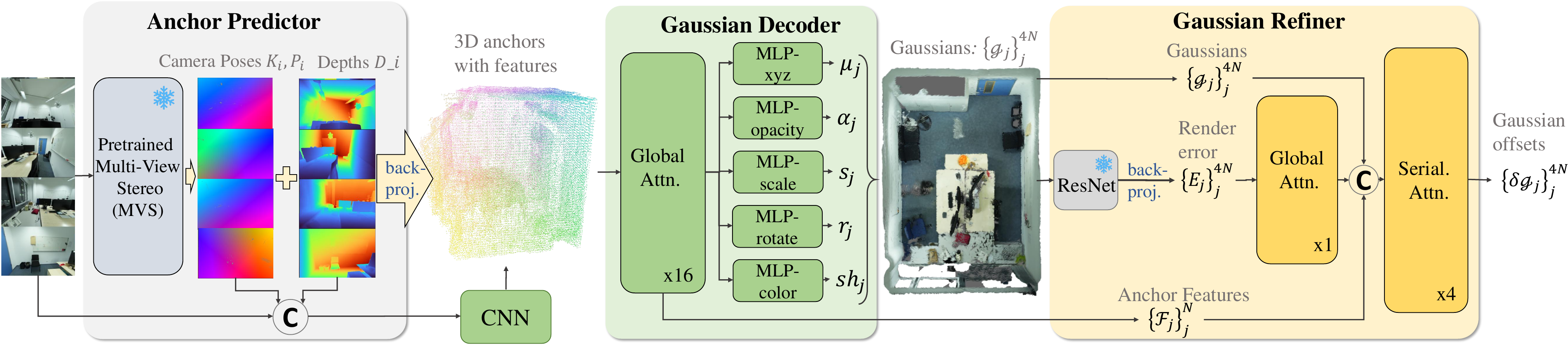}
  \caption{Overview of the proposed AnchorSplat pipeline.
The framework consists of three components: a pretrained Multi-View stereo module that extracts 3D geometric priors and depth maps from posed images, a transformer-based Gaussian decoder that predicts anchor-aligned Gaussians guided by these priors, and a lightweight Gaussian refiner that further refines Gaussian attributes to improve reconstruction quality and view consistency.}
  \label{fig:Pipeline}
  \vspace{-0.5cm}
\end{figure*}

%% file: sec/4_experiments.tex
\section{Experiment}
\label{sec:experiment}
\subsection{Experiment Setup}

\noindent\textbf{Implementation and Training.}  
We implement our model in PyTorch and leverage Ascend Flash Attention~\cite{liao2021ascend} for efficient attention computation. All computations are performed in bfloat16 precision.
The model is optimized using AdamW~\cite{loshchilov2017fixing}, and training is conducted on 64 Ascend 910B3 (64GB) NPUs. 
In the first stage, the Gaussian Decoder with 84M parameters is trained for 5k steps, followed by 5k steps for the Gaussian Refiner with 31M parameters. In detail, the Spherical Harmonics degree is set to 0. The Gaussian Decoder consists of 16 attention blocks with 640 channels and 2 single-layer MLP blocks to predict Gaussian attributes. The Gaussian Refiner comprises 1 attention block with 512 channels, 4 serialized attention blocks with 512 channels, and a single-layer MLP to refine existing Gaussian attributes. We will release our code, pre-trained models, and videos to facilitate reproducibility.

\noindent\textbf{Dataset.}  
To evaluate our method comprehensively, we conduct experiments on the ScanNet++ V2 dataset~\cite{yeshwanth2023scannet++} for generalizable 3D scene reconstruction. Input and novel views are uniformly sampled based on the view information provided in the dataset. Specifically, we select 32, 48, or 64 input views and 4, 6, or 8 novel views per scene. 
Input images are used at a resolution of $1168 \times 1752$, while rendering and supervision are performed at $448 \times 672$. 

\noindent\textbf{Baselines.}  
We adopt AnySplat~\cite{jiang2025anysplat} as our baseline, one of the few open-source voxel-aligned state-of-the-art methods. Pixel-aligned feed-forward approaches, as discussed earlier, are largely constrained by the number and resolution of input views. Therefore, we focus on comparisons with optimization-based methods and AnySplat.

\noindent\textbf{Evaluation Metrics.}
To evaluate novel view synthesis quality, we compute PSNR, SSIM, and LPIPS between predicted and ground-truth images. To assess the accuracy of predicted spatial geometry and multi-view consistency, we employ two widely used depth metrics: Absolute Relative Error (AbsRel) and $\delta_1$, defined as
\begin{align}
\text{AbsRel} &= \frac{1}{N} \sum_{i=1}^N \frac{| \hat D_i - D_i |}{D_i}, \nonumber \\
\delta_1 &: \max \left( \frac{\hat D_i}{D_i} , \frac{D_i}{\hat D_i} \right) < 1.25.
\end{align}
Additionally, we report the number of Gaussians and reconstruction time to evaluate the efficiency of our anchor-aligned Gaussian representation.

\input{tables/main_results}

\input{figures/3Dvis}

\input{tables/ablation_on_num_views}

\input{figures/anchorspalt_psnr_recontime}

\input{figures/refiner}

\input{tables/diff_views_very_sparse_dense}

\input{figures/mainResults}

\subsection{Experimental Results}
In the following results, AnchorSplat$^\star$ represents our AnchorSplat without using the Gaussian Refiner. 
As shown in Tab.~\ref{tab:main_results_4views}, we compare AnchorSplat with 3DGS~\cite{kerbl20233d}, Mip-Splatting~\cite{yu2024mip}, and AnySplat~\cite{jiang2025anysplat} to evaluate reconstruction performance. We choose these baselines for the following reasons. Many pixel-aligned feed-forward methods, as discussed earlier, are constrained by the number and resolution of input views, making them less suitable for scene-level reconstruction. We therefore adopt optimization-based methods as baselines in this setting. Among voxel-aligned approaches, AnySplat is one of the few open-source state-of-the-art methods, making it a particularly relevant comparison target. The results show that the optimization-based methods (3DGS and Mip-Splatting) achieve high-quality reconstructions on training views but generalize poorly to novel views, often exhibiting Gaussian artifacts, shadows, and floaters. AnySplat first predicts pixel-aligned Gaussians and then prunes redundant ones via differentiable voxelization, but it still requires a large number of Gaussians to represent the scene. Moreover, due to its pixel-aligned formulation, regions observed more frequently in the training views tend to generate more Gaussians, leading to artifacts and geometrically inconsistent predictions. Depth metric comparisons further support these observations. In contrast, our method uses significantly fewer Gaussians---nearly $1/20$ of those used by AnySplat---to represent the same scene, while achieving competitive or superior reconstruction quality with lower reconstruction time.

We further visualize the reconstructed 3D Gaussians produced by AnySplat and our method in Fig.~\ref{fig:3dvis}. AnySplat, despite being a voxel-aligned method, suffers from inaccurate geometry, resulting in floating artifacts, ghost structures, and irregular density distributions. In contrast, our anchor-aligned representation produces substantially cleaner and more coherent geometry with far fewer Gaussians. The reconstructed structures are sharper, more faithful to the scene geometry, and largely free of floaters and spurious points. 
We also compare the visualization of the rendered novel views by AnySplat and AnchorSplat, as shown in Fig.~\ref{fig:mainResult}.
These results demonstrate that our anchor-aligned design provides a more efficient and geometry-consistent representation, enabling stable and reliable 3D Gaussian reconstruction even under limited-view settings.

\subsection{Further Analyses}

\noindent\textbf{Quantitative Comparison with AnySplat using Different View Numbers.}
We further investigate the influence of the number of training views on reconstruction quality by experimenting with 32, 48, and 64 input views, and evaluating on 4, 6, and 8 novel views per scene, respectively. As shown in Tab.~\ref{tab:ablation_on_num_views} and Fig.~\ref{fig:psrn_recontime}, increasing the number of input views consistently improves reconstruction performance across all metrics, including PSNR, SSIM, and LPIPS. This demonstrates that richer multi-view information allows our model to better resolve occlusions, recover fine geometric details, and synthesize novel views more accurately.
Unlike voxel- or pixel-aligned methods such as AnySplat, where increasing the number of input views directly increases the number of predicted Gaussians and thus computational cost, our anchor-aligned Gaussians are fixed in number and determined by the spatial anchors. This design ensures that the representation remains stable and efficient regardless of input view count. Consequently, our method achieves high-quality novel view synthesis without a proportional increase in runtime or memory, highlighting the advantage of anchor-aligned Gaussians as a more compact and effective 3D scene representation. 


We further evaluate AnchorSplat under extremely sparse (3--5 views) and dense (128--256 views) input settings, as shown in Table~\ref{tab:ablation_on_very_views}. AnchorSplat consistently maintains stable performance while achieving efficient reconstruction with far fewer Gaussians.

\noindent\textbf{Visualisation Comparison of the effectvieness of the refiner module.}
We provide a visual comparison before and after applying the Gaussian Refiner to illustrate its effectiveness. As shown in the Fig.~\ref{fig:refiner}, the refiner significantly improves regions where the initial Gaussians predicted by the Gaussian Decoder struggle. After refinement, object boundaries become sharper and more accurate, missing regions and holes are filled in, and color consistency is noticeably improved. These qualitative results demonstrate that the Gaussian Refiner effectively corrects errors in the initial Gaussians and enhances the overall rendering quality.

%% file: tables/main_results.tex
\begin{table*}[ht!]
\centering
\caption{Novel 4-view evaluation on ScanNet++ with 32 input views. AnchorSplat achieves competitive or superior quality with nearly 20× fewer Gaussians and lower reconstruction time compared with 3DGS, Mip-Splatting, and AnySplat.}
\resizebox{0.86\textwidth}{!}{
\begin{tabular}{@{}lllccccccc@{}}
\toprule
\multirow{2}{*}{Dataset} & \multirow{2}{*}{Model} & \multirow{2}{*}{Category} & \multicolumn{3}{c}{RGB} & \multicolumn{2}{c}{Depth} & \multirow{2}{*}{NumGS} & \multirow{2}{*}{ReconTime(s)}\\
\cmidrule(lr){4-6} \cmidrule(lr){7-8} & & & \multicolumn{1}{c}{PSNR$\uparrow$} & \multicolumn{1}{c}{SSIM$\uparrow$} & \multicolumn{1}{c}{LPIPS$\downarrow$} & \multicolumn{1}{c}{$\delta_1\uparrow$} & \multicolumn{1}{c}{AbsRel$\downarrow$} & & \\
\midrule
\multirow{5}{*}{ScanNet++} & 3DGS                        &      Opt.  &  19.98                &   0.72                  &        \textbf{0.30}                    &          0.31             &       0.42                 &       496,087       &         391.44             \\
& Mip-Splatting                &    Opt.    &   19.92 &   0.75   &    0.34   &   0.35    &              0.38        &                398,212       &         289.95              \\
& AnySplat                    &    feed-forward   &     20.20  & 0.73 & 0.32 & 0.71 & 0.16 & 5,550,940 & 6.83 \\
& AnchorSplat$^\star$          &     feed-forward  & 20.96& 0.78& 0.47& \textbf{0.94}& 0.068& 247,153                       &                            \textbf{3.11} \\  
& AnchorSplat &     feed-forward  & \textbf{21.48} & \textbf{0.79} &                            0.38& \textbf{0.94} & \textbf{0.066} & \textbf{247,153} &                          5.52  \\ 
\bottomrule
\end{tabular}
}

\label{tab:main_results_4views}
\end{table*}

%% file: figures/3Dvis.tex
\begin{figure}[tp]
  \centering
   \includegraphics[width=0.9\linewidth, trim=10 20 14 10, clip]{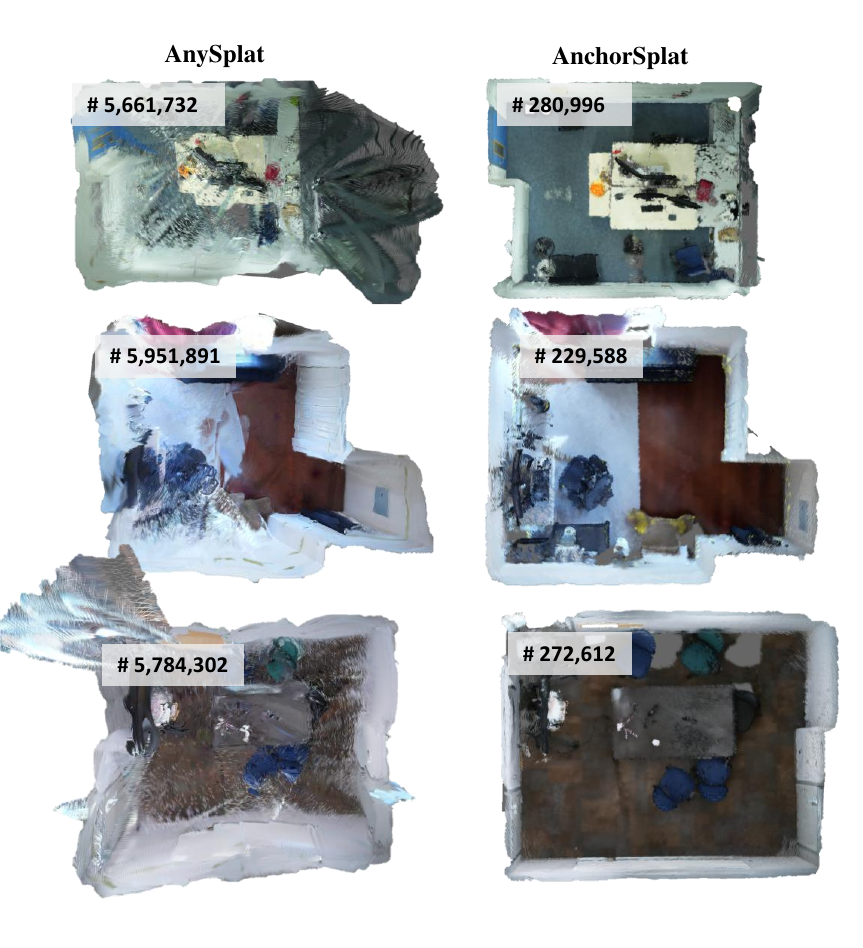}
   \caption{Comparison of reconstructed 3D Gaussians. Compared with AnySplat, our AnchorSplat produces cleaner and more geometry-consistent structures with far fewer Gaussians, reducing floaters, ghost artifacts, and irregular density distributions.}
   \label{fig:3dvis}
   \vspace{-0.3cm}
\end{figure}

%% file: tables/ablation_on_num_views.tex
\begin{table*}[ht!]
\centering
\caption{Evaluation on ScanNet++ V2 with 32/48/64 input views and 4/6/8 novel views, respectively. More input views improve reconstruction quality, while our anchor-aligned representation remains efficient with a fixed number of Gaussians.}
\resizebox{0.85\textwidth}{!}{
\begin{tabular}{@{}p{2.5cm}lccccccc@{}}
\toprule
\multirow{2}{*}{\centering Exp setting} & \multirow{2}{*}{Model} 
& \multicolumn{3}{c}{RGB} 
& \multicolumn{2}{c}{Depth} 
& \multirow{2}{*}{NumGS} & \multirow{2}{*}{ReconTime(s)}\\ 
\cmidrule(lr){3-5} \cmidrule(lr){6-7}
& & \multicolumn{1}{c}{PSNR$\uparrow$} & \multicolumn{1}{c}{SSIM$\uparrow$} & 
\multicolumn{1}{c}{LPIPS$\downarrow$} & \multicolumn{1}{c}{$\delta_1$$\uparrow$} & 
\multicolumn{1}{c}{AbsRel$\downarrow$} & & \\ 
\midrule
\multirow{3}{*}{32 sv / 4 nv} 
& AnySplat                    & 20.20  & 0.73 & \textbf{0.32} & 0.71 & 0.16 & 5,550,940 & 6.83 \\ 
& AnchorSplat$^\star$ & 20.96& 0.78& 0.47& \textbf{0.94} & 0.068& 247,153 &  \textbf{3.11}\\ 
& AnchorSplat                 & \textbf{21.48} & \textbf{0.79} &  0.38& \textbf{0.94} & \textbf{0.066} & \textbf{247,153} &  5.52\\ 
\midrule
\multirow{3}{*}{48 sv / 6 nv} 
& AnySplat                    & 20.66  & 0.74 & \textbf{0.31} & 0.67 & 0.17 & 8,197,441 & 7.57  \\ 
& AnchorSplat$^\star$ & 20.80& 0.78& 0.48& \textbf{0.94}& \textbf{0.064} & 247,153 & \textbf{3.23}\\ 
& AnchorSplat                 & \textbf{21.42} & \textbf{0.79} &  0.38& \textbf{0.94} & 0.066 & \textbf{247,153} & 5.43\\ 
\midrule
\multirow{3}{*}{64 sv / 8 nv} 
& AnySplat                    & 20.78 & 0.73 & \textbf{0.32} & 0.72 & 0.16 & 10,660,487 & 9.54 \\ 
& AnchorSplat$^\star$ & 21.10& 0.78& 0.47& \textbf{0.94} & \textbf{0.064} & 247,153 &  \textbf{3.71}\\ 
& AnchorSplat                 & \textbf{21.82} & \textbf{0.80} &  0.37& \textbf{0.94} & 0.065 & \textbf{247,153} &  6.13\\ 
\bottomrule
\end{tabular}
}
\label{tab:ablation_on_num_views}
\vspace{-2mm}
\end{table*}

%% file: figures/anchorspalt_psnr_recontime.tex
\begin{figure}[thbp]
  \centering
   \includegraphics[width=\linewidth, trim=6 2 2 2, clip]{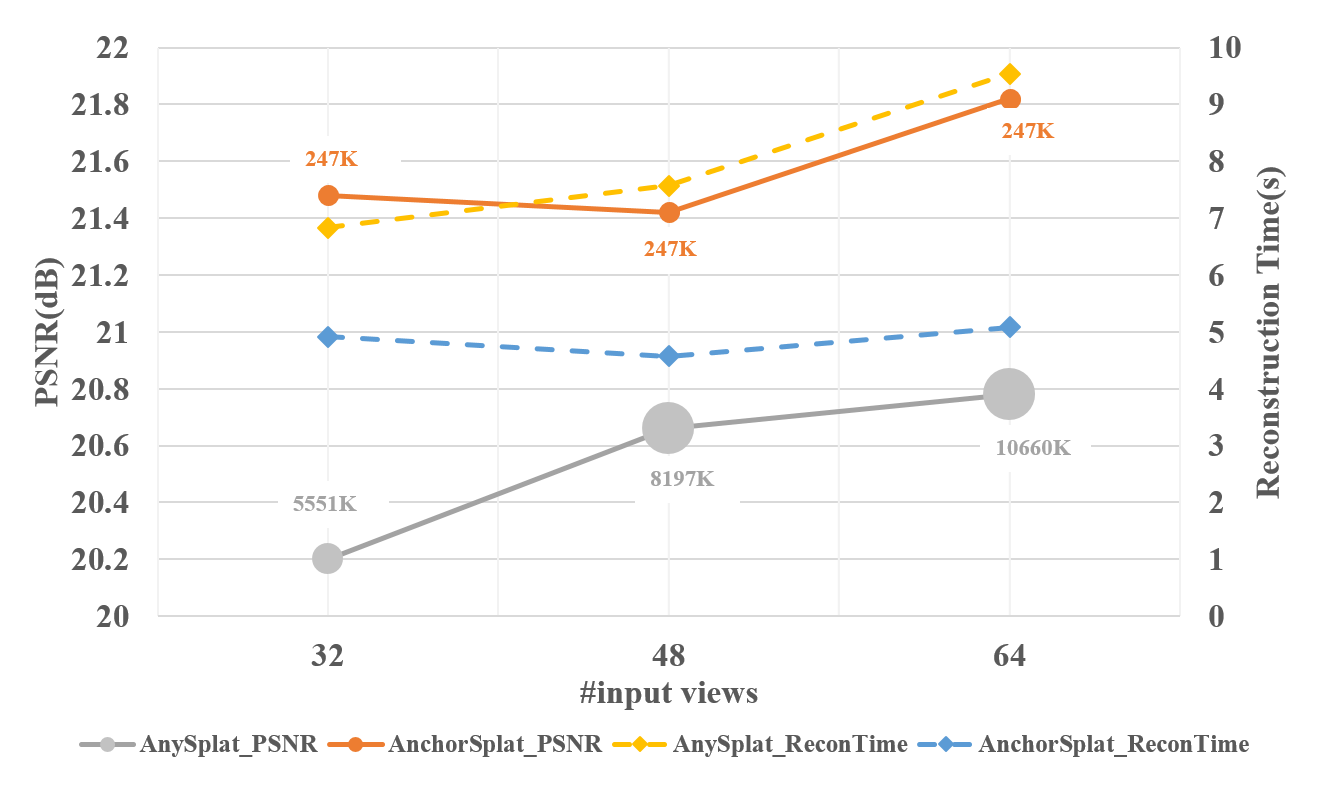}
    \caption{Reconstruction quality and runtime comparison between AnySplat and AnchorSplat across input-view settings. Dashed lines denote runtime, solid lines denote novel-view PSNR, and marker size with annotated numbers indicates the number of Gaussians (NumGS). AnchorSplat achieves better quality with lower runtime and fewer Gaussians.}
   \label{fig:psrn_recontime}
   \vspace{-0.3cm}
\end{figure}

%% file: figures/refiner.tex
\begin{figure}[tbp]
  \centering
   \includegraphics[width=0.9\linewidth, trim=6 5 6 0, clip]{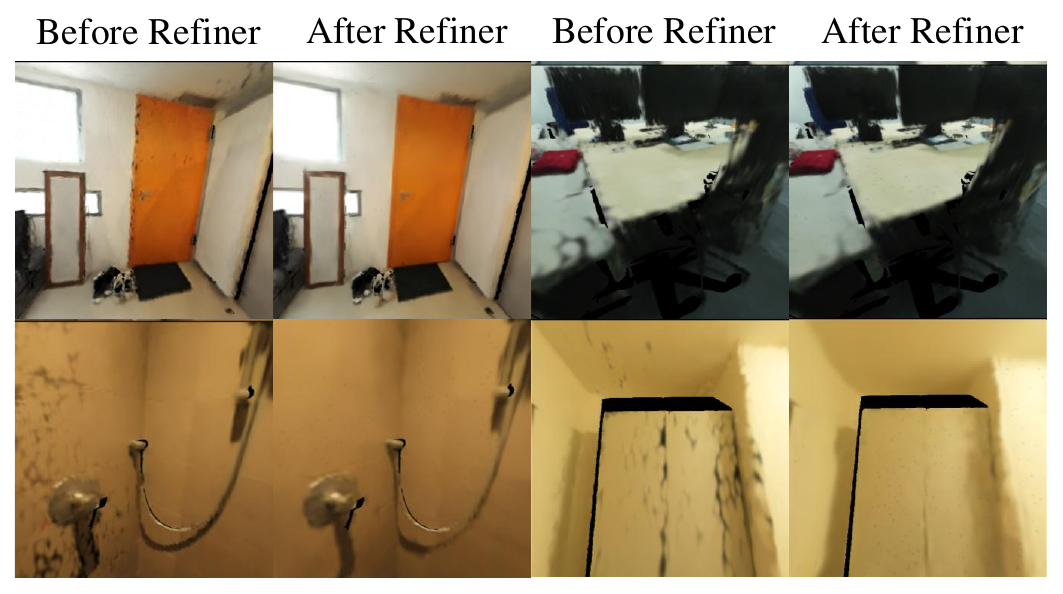}
   \caption{Visual comparison before and after applying the Gaussian Refiner. The refiner improves boundary sharpness, fills missing regions, and enhances color consistency, leading to better overall rendering quality.}
   \label{fig:refiner}
   \vspace{-0.5cm}
\end{figure}

%% file: tables/diff_views_very_sparse_dense.tex
\begin{table*}[ht]
\centering
\caption{Ablation results under extremely sparse (3/5) and extremely dense (128/256) input-view settings. AnchorSplat maintains stable performance while reconstructing scenes efficiently with far fewer Gaussians.}
\resizebox{0.86\textwidth}{!}{
\begin{tabular}{@{}llccccccc@{}}
\toprule
\multirow{2}{*}{Exp setting \quad \quad } & \multirow{2}{*}{Method \quad \quad \quad \quad \quad \quad  } & \multicolumn{3}{c}{RGB} & \multicolumn{2}{c}{Depth} & \multirow{2}{*}{NumGS\quad} & \multirow{2}{*}{ReconTime(s)} \\ \cmidrule(lr){3-5} \cmidrule(lr){6-7} 
sv/nv & & PSNR$\uparrow$ & SSIM$\uparrow$ & LPIPS$\downarrow$ & $\delta_1\uparrow$ & AbsRel$\downarrow$ & & \\ \midrule
\multirow{2}{*}{3 sv / 1 nv} & AnySplat    & 19.51 & 0.68 & \textbf{0.38} & 0.70 & 0.18 & 543,987 & \textbf{1.34 } \\
 & AnchorSplat$^\star$     & \textbf{19.99} & \textbf{0.78} & \textbf{0.38} & \textbf{0.92} & \textbf{0.073} & \textbf{247,153} & 3.18  \\ 
\multirow{2}{*}{5 sv / 1 nv} & AnySplat     & 20.21 & 0.77 & \textbf{0.36} & 0.71 & 0.17 & 909,132 & \textbf{2.25} \\
 &   AnchorSplat$^\star$    & \textbf{20.35} & \textbf{0.78} & 0.38 & \textbf{0.92} & \textbf{0.073} & \textbf{247,153} & 3.23  \\ 
 \cmidrule[0.1pt](lr){1-9}
\multirow{2}{*}{128 sv / 16 nv} & AnySplat   & 20.47 & 0.74 & \textbf{0.34} & 0.75 & 0.18 & 19,767,552 & 31.67   \\ 
 &  AnchorSplat$^\star$    & \textbf{21.58} & \textbf{0.79} & 0.37 & \textbf{0.94} & \textbf{0.058} & \textbf{247,153} & \textbf{7.36} \\ 
\multirow{2}{*}{256 sv / 32 nv} & AnySplat     & \multicolumn{7}{c}{OOM} \\
 &  AnchorSplat$^\star$  & \textbf{21.42} & \textbf{0.79} & \textbf{0.38} & \textbf{0.93} & \textbf{0.064} & \textbf{247,153} & \textbf{10.21} \\ 
\bottomrule
\end{tabular}
}
\label{tab:ablation_on_very_views}
\vspace{-3mm}
\end{table*}

%% file: figures/mainResults.tex
\begin{figure*}[thbp]
  \centering
  \includegraphics[width=0.85\linewidth]{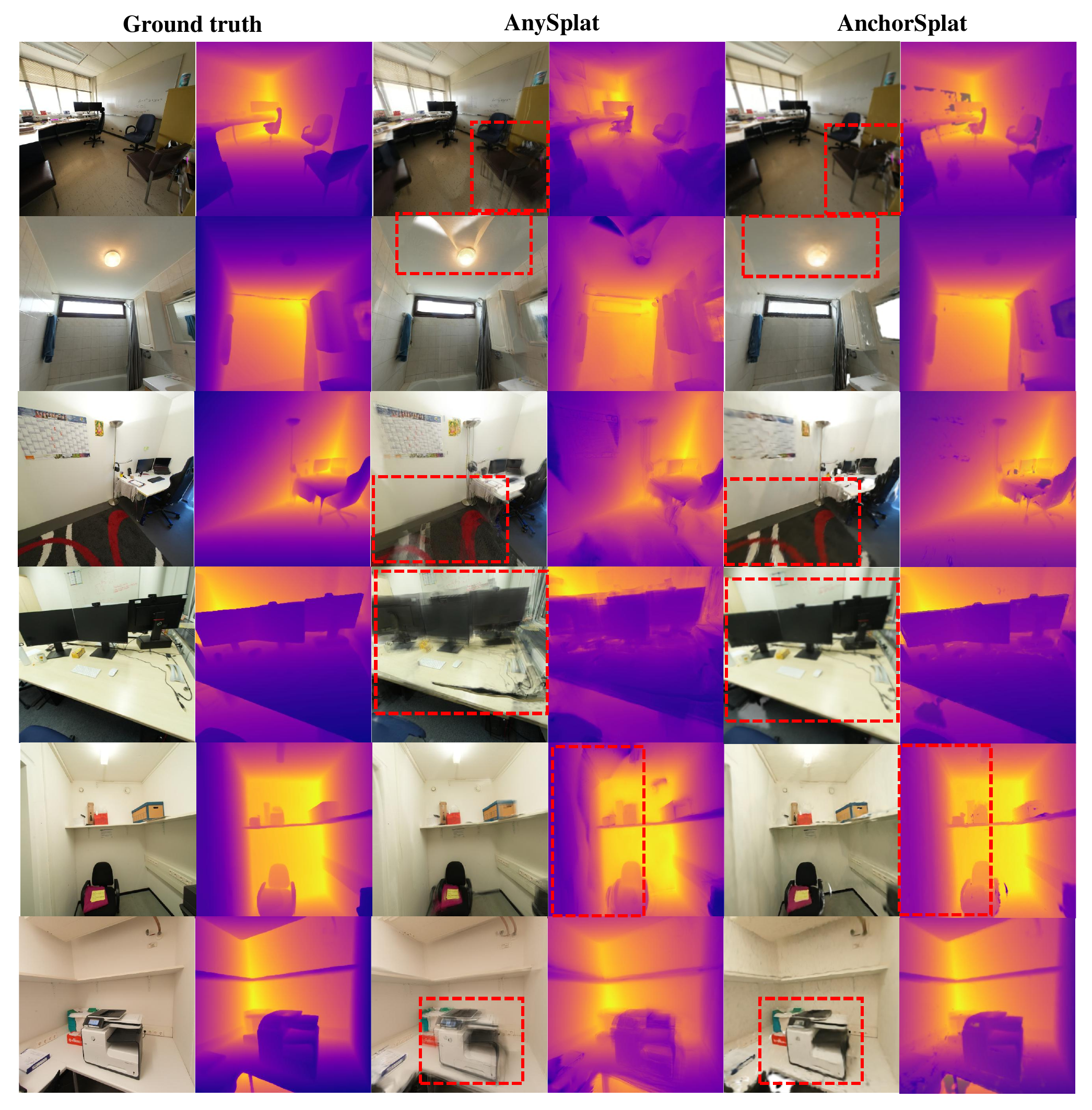}
  \caption{Comparison visualization. AnchorSplat produces noticeably higher-quality renderings with more accurate geometry and sharper depth, while avoiding the multi-view inconsistencies that often lead to artifacts such as ghosting and floaters. For example, in the first row, the chair region rendered by AnySplat exhibits clear view-dependent artifacts caused by inconsistent multi-view alignment, whereas our method maintains clean geometry and coherent appearance. Similar improvements can be observed across various scenes in the visualization, highlighting the robustness and reliability of our anchor-aligned Gaussian representation.}
  \label{fig:mainResult}
  \vspace{-0.1cm}
\end{figure*}

%% file: sec/5_conclusion.tex
\section{Conclusion}
\label{sec:conclusion}

In this work, we presented AnchorSplat, a feed-forward 3D Gaussian Splatting framework for scene-level reconstruction in native 3D space. By introducing anchor-aligned Gaussians guided by geometric priors, our method decouples the representation from input image resolution and viewpoint density, enabling a more geometry-aware and efficient scene representation. Our two-stage design, with a Gaussian decoder followed by a Gaussian refiner, further improves reconstruction quality and cross-view consistency. Experiments on the ScanNet++ V2 benchmark demonstrate that AnchorSplat achieves highly plausible and view-consistent 3D reconstructions with fewer Gaussians and lower computational cost than pixel-aligned baselines.
%
%
\noindent\textbf{Limitations.}
Despite its effectiveness, AnchorSplat depends on reasonably accurate geometric priors. When these priors are incomplete, constrained Gaussian growth and budget make empty regions difficult to cover, which may degrade reconstruction quality. Future work will explore adaptive density control and dynamic Gaussian growth.



%% file: sec/X_suppl.tex
\clearpage
\setcounter{page}{1}
\maketitlesupplementary

\appendix

\section{Appendix}
This supplementary material provides the following additional information: (B) additional ablation experiments and results and (C) additional visualizations.

\section{Ablation Study}
\label{sup_sec:ablation}

\subsection{Ablation Study on Input Information}
We perform an ablation study on the ScanNet++ V2 dataset to investigate how different input modalities influence the performance of the GS decoder. The following configurations are examined: (i) RGB-only, (ii) RGB with camera ray embeddings, (iii) RGB with depth, and (iv) RGB supplemented with both camera information and depth. To isolate the effect of input modalities, this study evaluates the Gaussian decoder alone without any refinement, and thus only \textbf{Ours (w/o Refiner)} is reported. As summarized in Table.~\ref{tab:sup_abla_input}, incorporating additional geometric and camera-related cues leads to progressively better reconstruction quality, demonstrating the importance of informed inputs for effective Gaussian decoding.


\input{tables/suppl_abla_inputInfo}

\subsection{Ablation Study on number of Gaussians}

We further conduct an ablation study to analyze the effect of the number of Gaussians predicted per anchor in the Gaussian decoder on the ScanNet++ V2 dataset. Specifically, we vary the Gaussian count per anchor across {1, 2, 4, 8, 16} and evaluate how this design choice influences reconstruction fidelity and geometric accuracy, as shown in Table.~\ref{tab:sup_abla_numGS}. To isolate the capacity of the decoder itself, this experiment evaluates only \textbf{Ours (w/o Refiner)}, without applying the Gaussian refiner.

The results show that the number of Gaussians per anchor plays a critical role in balancing expressiveness and stability. Very small configurations lack sufficient representational capacity, leading to incomplete or under-detailed geometry. In contrast, overly large configurations yield only marginal accuracy gains while incurring significantly higher computational cost. In practice, we therefore adopt 4 Gaussians per anchor as the default configuration, as it achieves the best overall performance while keeping computational cost low.
This ablation study is conducted using a smaller batch size (bs = 32) while keeping the same number of training iterations (2.5k) as in the main experiments. As a result, the absolute performance values may differ from those in the main paper, but the overall trend regarding Gaussian multiplicity remains consistent. Moreover, when incorporating our Gaussian Refiner, the reconstruction quality surpasses even the decoder-only configuration with 16 Gaussians per anchor, further demonstrating the effectiveness of the refinement module.



\input{tables/suppl_abla_numGS}

\subsection{Ablation Study on Variant Datasets}
We also conducted ablation experiments on additional datasets, including indoor datasets (Replica) and the outdoor dataset Tanks and Temples (T\&T), as shown in Table~\ref{tab:abla_datasets}. Here, AnchorSplat$^\star$ denotes AnchorSplat without the Refiner. The results demonstrate that our method achieves superior RGB and depth rendering quality, as well as higher reconstruction efficiency, compared to AnySplat. These results further validate the generalization ability and robustness of our approach across diverse scenes. 

\input{tables/ablation_diff_datasets}

\subsection{Ablation Study on Aggregation Method}
For each anchor, multiple projected image features may be obtained from different views that map to the same anchor. We aggregate these features into a single $C$-dimensional anchor feature by applying a pooling operation over all valid projections that pass the visibility and depth-consistency checks. We conducted an early-stage ablation study to compare different aggregation strategies, including average pooling, max pooling, and a FIFO (first in fist out) selection baseline.
We first visualized the aggregated features using PCA, as shown in Fig.~\ref{fig:feature_aggreg_vis}, and then evaluated the three pooling methods quantitatively. The results indicate PSNR values of 20.96, 20.81, and 20.28 for average, max, and FIFO pooling, respectively. Based on these results, we adopt average pooling as the default aggregation method.

 For each anchor, we may obtain multiple projected image features (from different views mapping to the same anchor). We aggregate these features into a single  C-dimensional anchor feature by applying a pooling operation over all valid projected features associated with that anchor, and we only pool over projections that pass the visibility/depth-consistency checks.
We conducted an early-stage ablation comparing different aggregation strategies, including average pooling, max pooling, and a FIFO/first-in selection baseline.
 We initially performe PCA visualization shown as Fig.~\ref{fig:feature_aggreg_vis} and experiments on features aggregated using three pooling methods (average, max, FIFO). The results showed PSNR values of 20.96, 20.81, and 20.28 for average, max, and FIFO pooling, respectively. Therefore, we utilize the average pooling as default method.

\input{figures/vis_feature_aggreg}

\input{figures/supl_3dvis}

\input{figures/supl_2dvis}


\section{Ablation Study on Backbones}
In the AnchorPredictor module, we use MapAnything as the default backbone for depth prediction. This backbone estimates the depth at each anchor location, providing a strong geometric prior for the subsequent AnchorSplat reconstruction. To evaluate the impact of the backbone choice, we compare MapAnything with DA3 as the AnchorPredictor backbone on the Replica dataset. As shown in Table~\ref{tab:abla_backbones}, both backbones achieve comparable performance given the same number of training steps, demonstrating that MapAnything serves as an effective and reliable default for depth prediction.

\input{tables/ablation_backbones}

\section{Additional Visualizations}
\label{sup_sec:add_vis}

When converting the depth maps produced by the MVS estimator into 3D points via back-projection, we observe that some predicted depths are unreliable, resulting in outlier 3D points such as flying points, points lying far behind actual surfaces, or points drifting outside the valid scene region. To address these artifacts, we apply a 3D clipping operation that restricts all back-projected points to a predefined spatial boundary. As illustrated in Fig.~\ref{fig:supl_3dvis}, this clipping step effectively removes extreme outliers, stabilizes the initial geometry, and ensures that the Gaussian initialization remains structurally valid without being affected by large-magnitude depth errors.

In addition, we provide extended visual comparisons against AnySplat to more clearly demonstrate the advantages of our anchor-aligned representation. As shown in Fig.~\ref{fig:supl_2dvis}, both the RGB and depth rendering results reveal that our method consistently produces sharper appearance details and significantly cleaner geometry, free from the floaters, ghosting artifacts, and structural distortions commonly observed in voxel-aligned approaches. The depth visualizations are particularly indicative of this improvement: our predictions preserve crisp geometric boundaries and exhibit strong view-consistency, suggesting that the underlying 3D structure is substantially more accurate and stable.

These observations are further corroborated by the reconstructed Gaussian visualizations in Fig.~\ref{fig:supl_3dvis}. Our method generates compact, well-structured, and spatially coherent Gaussian distributions, whereas AnySplat tends to produce fragmented, noisy, and geometrically inconsistent Gaussians. Together, these comparisons highlight that AnchorSplat not only improves rendering quality but also delivers a more faithful and robust 3D representation.

%% file: tables/suppl_abla_inputInfo.tex
\begin{table}[ht!]
\centering
\caption{Ablation study on input information for the Gaussian decoder. This experiment presents the evaluation results on the novel 4 views using ScanNet++ V2 with 32 input views. In this table, RGB denotes using only RGB images as input; Depth indicates the use of depth maps; CamRay represents Plücker ray embeddings derived from camera intrinsics and extrinsics.}
\resizebox{\columnwidth}{!}{
\begin{tabular}{@{}p{3.5cm}cccccc@{}}
\toprule
\multirow{2}{*}{Setting} & \multicolumn{3}{c}{RGB}                                                            & \multicolumn{2}{c}{Depth}                                 \\ \cmidrule(l){2-4} \cmidrule(l){5-6} 
                         & \multicolumn{1}{c}{PSNR$\uparrow$} & \multicolumn{1}{c}{SSIM$\uparrow$} & \multicolumn{1}{c}{LPIPS$\downarrow$} & \multicolumn{1}{c}{$\delta_1\uparrow$} & \multicolumn{1}{c}{AbsRel$\downarrow$}  \\ \midrule
RGB                      &                           20.53&                           0.78&                            0.46&                             0.92&                             0.076\\
RGB+Depth                &                           20.67&                           0.78&                            0.48&                             0.94&                             0.057\\
RGB+CamRay               &                           20.34&                           0.78&                            0.48&                             0.92&                             0.080\\
RGB+CamRay+Depth  & 20.96 & 0.78 &  0.47 & 0.94 & 0.068  \\ \bottomrule
\end{tabular}
}
\label{tab:sup_abla_input}
\end{table}

%% file: tables/suppl_abla_numGS.tex
\begin{table}[ht!]
\centering
\caption{Ablation on the number of Gaussians per anchor.
We evaluate the effect of varying the Gaussian multiplicity in the Gaussian decoder on the ScanNet++ V2 dataset. All results are obtained using the decoder only (w/o Refiner). Using too few Gaussians restricts representational capacity, while very large counts yield diminishing returns with substantially higher computational cost. A moderate setting of 4 Gaussians per anchor achieves the best overall balance between accuracy and efficiency.}
\resizebox{\columnwidth}{!}{
\begin{tabular}{@{}lcccccc@{}}
\toprule
\multirow{2}{*}{Setting} & \multicolumn{3}{c}{RGB}                                                            & \multicolumn{2}{c}{Depth}    & \multirow{2}{*}{numGS}  \\ \cmidrule(l){2-4} \cmidrule(l){5-6} 
                         & \multicolumn{1}{c}{PSNR$\uparrow$} & \multicolumn{1}{c}{SSIM$\uparrow$} & \multicolumn{1}{c}{LPIPS$\downarrow$} & \multicolumn{1}{c}{$\delta_1\uparrow$} & \multicolumn{1}{c}{AbsRel$\downarrow$}  \\ \midrule
numGS=1        &       20.14               &       0.78               &       0.50               &       0.93               &          0.077                  &               61,788         \\
numGS=2                  &      20.61          &        0.79              &      { 0.48 }              &       0.93               &         0.075                   &            123,577            \\
numGS=4                 &       20.66               &       { 0.79 }             &        {0.48 }             &     0.93                 &          0.076                  &           247153             \\
numGS=8                  &         20.46             &          0.78            &          0.49            &    0.93                  &         0.073                   &          494,307              \\
numGS=16           & {20.79} & {0.79} & 0.49 & {0.94} &     {0.072}                &         988,613              \\ \bottomrule
\end{tabular}
}
\label{tab:sup_abla_numGS}
\end{table}

%% file: tables/ablation_diff_datasets.tex

\begin{table}[htbp]
\centering
\caption{Comparison of AnchorSplat variants on different datasets (32 sampled views / 4 novel views). Experiments are conducted on indoor datasets (ARkitScenes and Replica) and an outdoor dataset (T\&T), using the same number of training steps, learning rate, and backbone settings. AnchorSplat$^\star$ denotes AnchorSplat without the Refiner. Performance is evaluated in terms of RGB and depth rendering quality as well as reconstruction efficiency.}
\resizebox{0.5\textwidth}{!}{
\begin{tabular}{@{}llccccccc@{}}
\toprule
\multirow{2}{*}{Dataset} & \multirow{2}{*}{Method} & \multicolumn{3}{c}{RGB} & \multicolumn{2}{c}{Depth} & \multirow{2}{*}{NumGS} & \multirow{2}{*}{ReconTime(s)} \\ \cmidrule(lr){3-5} \cmidrule(lr){6-7} 
 & & PSNR$\uparrow$ & SSIM$\uparrow$ & LPIPS$\downarrow$ & $\delta_1\uparrow$ & AbsRel$\downarrow$ & & \\ \midrule
\multirow{2}{*}{ARKitScenes} & AnySplat    & 21.35 & 0.75 & 0.30 & 0.88 & 0.12 & 3,237,113 & 6.63 \\
 &  AnchorSplat$^\star$   & 21.00 & 0.77 & 0.41 & 0.96 & 0.06 & 400,000 & 2.23 \\
 \cmidrule[0.1pt](lr){1-9}
\multirow{2}{*}{Replica} & AnySplat   & 21.19& 0.70& 0.32& 0.86& 0.11& 5,740,107& 4.72   \\
  &  AnchorSplat$^\star$    & 23.48 & 0.78 & 0.31 & 0.95 & 0.066 & 800,000 & 2.40\\
  \cmidrule[0.1pt](lr){1-9}
   \multirow{2}{*}{T\&T} & AnySplat     & 15.30 & 0.45 & 0.46 & 0.44 & 0.39 & 3,952,092 & 6.74 \\
 &   AnchorSplat$^\star$    & 16.53 & 0.57 & 0.59 & 0.75 & 0.23 & 800,000 & 1.98 \\
\bottomrule
\end{tabular}
}
\label{tab:abla_datasets}
\end{table}

%% file: figures/vis_feature_aggreg.tex
\begin{figure}[thbp]
  \centering
   \includegraphics[width=\linewidth, trim=6 8 6 9, clip]{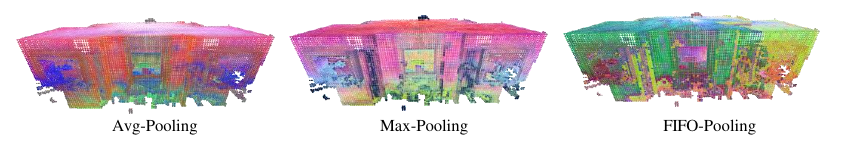}
   \caption{PCA visualization of three feature aggregations.}
   \label{fig:feature_aggreg_vis}
\end{figure}

%% file: figures/supl_3dvis.tex
\begin{figure}[thbp]
  \centering
   \includegraphics[width=\linewidth, trim=0 0 0 0, clip]{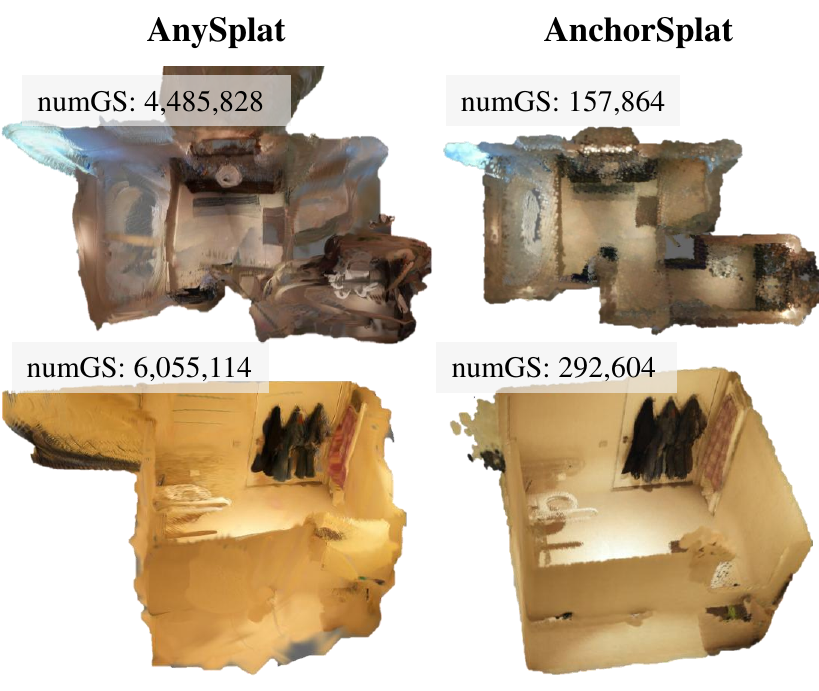}
   \caption{Comparison of reconstructed Gaussians between AnySplat and AnchorSplat}
   \label{fig:supl_3dvis}
\end{figure}

%% file: figures/supl_2dvis.tex
\begin{figure*}[tb]
  \centering
   \includegraphics[width=\linewidth, trim=0 0 0 0, clip]{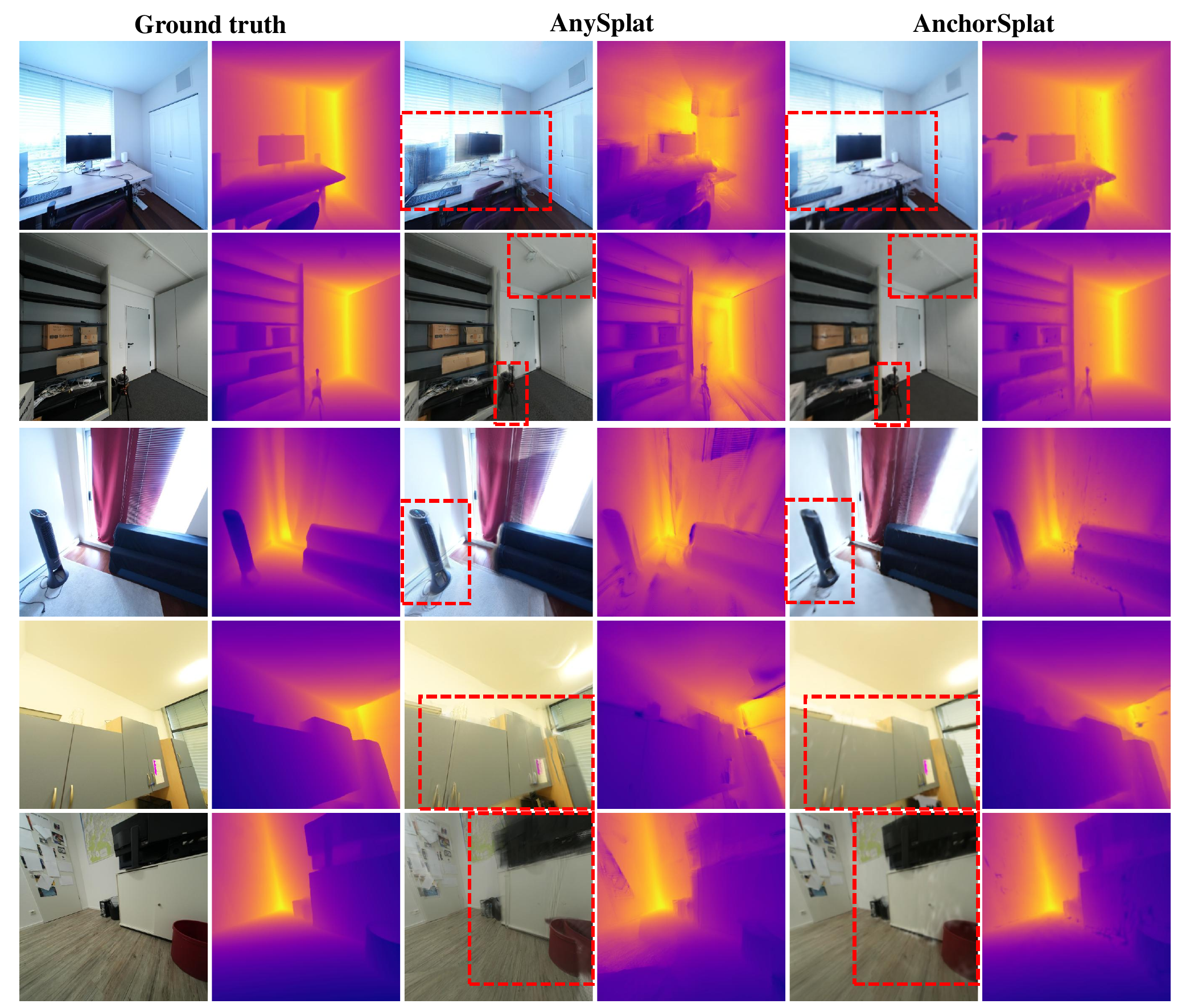}
   \caption{Comparison of rendered RGB images and depth images between AnySplat and AnchorSplat}
   \label{fig:supl_2dvis}
\end{figure*}

%% file: tables/ablation_backbones.tex
\begin{table}[htbp]
\centering
\caption{Ablation study of AnchorPredictor backbones on the Replica dataset (32 sampled views / 32 novel views). We compare MapAnything and DA3 as backbones under the same training configurations, including identical number of training steps, learning rate, and network settings. Performance is measured in terms of RGB and depth rendering quality as well as reconstruction efficiency.}
\resizebox{0.5\textwidth}{!}{
\begin{tabular}{@{}lccccccc@{}}
\toprule
\multirow{2}{*}{Backbone} & \multicolumn{3}{c}{RGB} & \multicolumn{2}{c}{Depth} & \multirow{2}{*}{NumGS} & \multirow{2}{*}{ReconTime(s)}\\ \cmidrule(lr){2-4} \cmidrule(lr){5-6} 
 & \multicolumn{1}{c}{PSNR$\uparrow$} & \multicolumn{1}{c}{SSIM$\uparrow$} & \multicolumn{1}{c}{LPIPS$\downarrow$} & \multicolumn{1}{c}{$\delta_1\uparrow$} & \multicolumn{1}{c}{AbsRel$\downarrow$} & & \\ \midrule
MapAnything & 22.41 & 0.79 & 0.33 & 0.92 & 0.084 & 800,000 & 1.89 \\
DA3       & 22.04 & 0.75 & 0.35 & 0.91 & 0.085 & 800,000 & 1.72 \\
\bottomrule
\end{tabular}
}
\label{tab:abla_backbones}
\end{table}